\documentclass[11pt,a4paper]{article}
\usepackage[hyperref]{emnlp2020}
\usepackage{times}
\usepackage{latexsym}

\usepackage{microtype}
\usepackage{times}
\usepackage{xcolor}
\usepackage{natbib}
\usepackage{soul}
\usepackage{tikz}
\usepackage{tikz-dependency}
\usetikzlibrary{positioning}
\usepackage[utf8]{inputenc}

\usepackage[nocenter]{qtree}

\usepackage{CJKutf8}
\usepackage{url}
\usepackage{tipa}

\usepackage{amsmath,amsthm,amssymb,amsfonts,bbm,stmaryrd}
\usepackage{enumitem}

\usepackage{textcomp}
\usepackage{multirow} % Allows what
\usepackage{graphicx} % Allows including images
\usepackage{subcaption}
\usepackage{booktabs} % Allows the use of \toprule, \midrule and 
\usepackage[normalem]{ulem}
\usepackage{makecell}
\usepackage{courier}
\usepackage{bbm}
\usepackage{algorithm,algpseudocode}
\usepackage{adjustbox}
\usepackage{tabularx}
\usepackage{booktabs}
\usepackage{setspace}

\usepackage{threeparttable}

\usepackage{todonotes}

\usepackage{listings}
\usepackage{pifont}
\usepackage{color}
\usepackage{bm}
\usepackage{ulem}
\usepackage{array}

\DeclareMathOperator*{\argmax}{argmax}

\usepackage{pifont}
\newcommand{\mbs}{\boldsymbol}

\newcommand{\mb}{\mathbf}

\newcommand{\bc}{\mathbf{c}}

\newcommand{\bw}{\mathbf{w}}

\newcommand{\bv}{\mathbf{v}}

\newcommand{\bz}{\mathbf{z}}
\newcommand{\bu}{\mathbf{u}}
\newcommand{\bmu}{\bm{\mu}}

\newcommand{\Ra}{\shortrightarrow}

\usepackage{siunitx}
\usepackage{ragged2e}
\newcolumntype{R}[1]{>{\RaggedLeft\arraybackslash}p{#1}}

\newcolumntype{P}[1]{>{\RaggedRight\arraybackslash}p{#1}}

\aclfinalcopy % Uncomment this line for the final submission
\title{Visually Grounded Compound PCFGs}

\newcommand{\uoa}{\normalfont \text{\textipa{\ae}}}
\newcommand{\uoe}{\normalfont \text{\textipa{E}}}

\author{Yanpeng Zhao$^{\uoe}$ \\
  $^{\uoe}$ILCC, University of Edinburgh \\
  \tt{yanp.zhao@ed.ac.uk} \\\And
  Ivan Titov$^{\uoe\uoa}$\\
  $^{\uoa}$ILLC, University of Amsterdam \\
  \tt{ititov@inf.ed.ac.uk} \\}

\date{}

\begin{document}
\maketitle
\begin{abstract}
%\sout{Natural language is often grounded in perceptual experiences.} 
%\Ivan{Not sure we need this first sentence. Drop?}\Reply{seems unnecessary to me}
%typically in visual images.
Exploiting visual groundings for language understanding has recently been drawing much attention.
In this work, we study visually grounded grammar induction
and learn a constituency parser from both unlabeled text and its visual groundings.
Existing work on this task~\citep{shi-etal-2019-visually} optimizes a parser via \textsc{Reinforce} and derives the learning signal only from the alignment of images and sentences.
While their model is relatively accurate overall, its error distribution is very uneven, with low performance on certain constituents types (e.g., 26.2\% recall on verb phrases, VPs) and high on others (e.g., 79.6\% recall on noun phrases, NPs). This is not surprising as the learning signal is likely insufficient for deriving all aspects of phrase-structure syntax and gradient estimates are noisy.  We show that using %a variant 
an extension of probabilistic context-free grammar  model we can do fully-differentiable end-to-end visually grounded learning.
Additionally, this enables us to complement the image-text alignment loss with a language modeling objective. 
%can as well derive learning signals from text counterpart,
%which complements the visually grounded learning in the sense that not all semantic units in a sentence are necessarily grounded in or aligned to an image.
On the %automatically parsed MSCOCO test captions,
%\Ivan{Do we have to emphasize this?} 
%\Reply{No.}
MSCOCO test captions,
our model establishes a new state of the art, outperforming its non-grounded version  and, thus,
confirming the effectiveness of visual groundings in constituency grammar induction. It also substantially outperforms
the previous grounded model, with largest improvements on more `abstract' categories (e.g., +55.1\% recall on VPs).\footnote{Our code is available at \href{https://github.com/zhaoyanpeng/vpcfg}{https://git.io/JU0JJ}.}
\end{abstract}

\section{Introduction}

% grammar induction
Grammar induction is a task of finding latent hierarchical structure of language.
%A tree structure consists of a set of hierarchically arranged rules, 
%i.e., \textit{syntax},
%describing the generation of language~\citep{chomsky_1980}. I: sorry, I'd rather not say that the rule describes the generation of language (I do not think it is a modern cognitively plausible view, maybe you mean generation in a different sense
%Inducing tree-structured representation is of paramount importance for language understanding tasks~\citep{chen-etal-2017-improved,he-etal-2017-deep}. \Ivan{I really don't want to say this. I don't think people believe that this is true. I certainly don't. }
As a fundamental problem in computational linguistics, % long-standing = for decades, i.e. tautology
it has been extensively studied for decades~\citep{LARI199035, carroll1992two,clark-2001-unsupervised,klein-manning-2002-generative}.
Recently, %neural-based 
deep learning models %\Ivan{`Neural-based deep learning' - tautololgical, choose one}
have been shown very effective across NLP tasks and
%in language understanding. 
have also been applied to grammar induction, greatly advancing the area~\citep{shen2018neural,shen2018ordered,kim-etal-2019-compound,kim-etal-2019-unsupervised,jin-etal-2019-unsupervised}.
%Despite impressive progress made by these approaches, 
These neural grammar-induction approaches have been generally limited to relying on text, 
without considering learning signals from other  modalities.
%\Ivan{\textbf{New comment: } I think you should just not discuss it here in the introduction. You can mostly they this is a well studied problem, and recent advance have been brought up with neural methods. However, the challenges still remain. Then you go into the second section about visiual grounding.}

% why visually grounded learning
%On the other hand, - you cannot have a connective 'on the other hand' without have 'on the one hand'
%In contrast,
%natural language %has been shown being grounded 
%learning is grounded
%in perceptual experiences
In contrast, the crucial aspect of natural language learning is that it is grounded in perceptual experiences
~\citep{barsalou_1999,fincher-2001,bisk2020experience}.
%\Ivan{Is this what you meant to say? You cannot say that this has been shown, I think. Not sure this even requires citations (so maybe drop). If you want to have citations, it may make sense to say instead 'In contrast, the crucial aspect of natural language learning is that it is grounded in perceptual experiences'}
%\Reply{Yes. I would prefer to keep the citations. I edited the sentence as you suggested.}
We thus anticipate improved language understanding by leveraging grounded learning.
Promising results from grounded learning  have been emerging in areas such as representation learning~\citep{bruni-2014,kiela-etal-2018-learning,bordes-etal-2019-incorporating}. %and word discovery~\citep{kawakami-etal-2019-learning}.
Typically, they use visual images as perceptual groundings of language and aim at improving  \textit{continuous vector representations} of language (e.g., word or sentence embeddings).
%As visual images may ground compositions of textual semantics,
%can we exploit visually grounded semantic compositions to induce tree-structured representation of language?
%Take the sentence `\textit{the sky above is blue}',
%it is commonly accepted that `\textit{above}' in the sentence describes a spatial relation in physical world rather than being a noun modified by `\textit{sky}'. 
%The conception of the \textit{spatial relation} semantic of `\textit{above}' is shaped by our perception of the world.
%We can thus anticipate that gounded learning can benefit language understanding. 
%\Ivan{Sorry, do not quite get the story in the last 3 sentences. Can we have an example which is more related to what is happening in our model, in that way we can link to it in the following text. }
%\Reply{This example can explain (1) understanding semantics of a word relying on perceptual experiences; (2) matching-based visually grounded learning may find constituents like `(the (sky above))'. Language modeling-based approaches are likely to find `((the sky) above)'. Maybe it is not a good example to show my points.}
% problem definition and previous work
In this work, we consider a more challenging problem:
can visual groundings help us induce \textit{syntactic structure}? 
%\Ivan{Emphasized syntax / polished (structure somehow reads better than trees...}
We refer to this problem as \textit{visually grounded grammar induction}.

\citet{shi-etal-2019-visually} propose a visually grounded neural syntax learner (VG-NSL) to tackle the task.
Specifically, they learn a parser from aligned image-sentence pairs (e.g., image-caption data),
where each sentence describes visual content of the corresponding image.
The parser is optimized via \textsc{Reinforce}, where the reward is computed by scoring the alignment of images and constituents.
While straightforward, matching-based rewards can, as we will discuss further in the paper, make the parser focus only on more local and short constituents (e.g., 79.6\% recall on NPs) and to perform poorly on longer ones (e.g., 26.2\% recall on VPs)~\citep{shi-etal-2019-visually}. 
%\Ivan{recall vs F1, numbers -- need to be %consistent across abstract / into. Just a note to %keep it in mind.}
%\Reply{Thanks. I noticed this issue. I will use %recall, as we can not compute per-label F1 in the %unsupervised setting}\Ivan{Ah! Makes sense}
While for the former it outperforms the text-only
grammar induction methods, for the latter it substantially underachieves.  
%\Ivan{Added to emphasize complementarity, makes sense / correct?}
%\Reply{fine to me.}
This may not be surprising, as it is not guaranteed that every constituent of a sentence has its visual representation in the aligned image;
the reward signals can be noisy and insufficient to capture all aspects of phrase-structure syntax.
Consequently, \citet{shi-etal-2019-visually} have to rely on language-specific inductive bias to obtain more informative reward signals.
%\Ivan{Maybe we should emphasize that grounding signal is likely insufficient to induce the structure of language? Or this is a too general statement, maybe we should make it about their objective? Would it make sense to clarify in a sentence what they are trying to do?}
Another issue with VG-NSL is that the parser does not admit tractable estimation of the partition function and the posterior probabilities for constituent boundaries needed to compute the expected reward in closed form. 
%\Ivan{I added about posteriors, as the one doesn't necessary imply the other. Felt slightly odd.}
%\Reply{Yes. It feels a bit odd. I would prefer to simply use "the partition function". It has a unambiguous indication in the parsing area? My understanding is that we can compute marginals (conditional probabilities here) as long as we can compute the partition function, is this correct?}
Instead, VG-NSL relies on Monte Carlo policy gradients, 
%and hence the expected rewards have to be computed via point-estimation --  \Ivan{Not sure I like the term 'point estimation'} 
potentially suffering from high variance. %\Ivan{Rephrased slightly}
%These two practical issues inhibit visually grounded learning of syntax.

% our solutions
To alleviate the first issue, 
we propose to complement the image-text alignment-based loss with a loss defined on unlabeled text (i.e., its log-likelihood).
%, rather than only from the alignments of constituents and images.
As re-confirmed with neural models  in~\citet{shen2018ordered} and~\citet{kim-etal-2019-compound},
text itself can drive induction of rich syntactic knowledge, so
additionally optimizing the parser on raw text can  be beneficial and complementary to  visual grounded learning.
To resolve the second issue,
we resort to an extension of probabilistic context-free grammar (PCFG) parsing model,  compound PCFG~\cite{kim-etal-2019-compound}. 
%\Ivan{Maybe an extension? Strictly, speaking it is not PCFG. Also, in abstract where you refer to C PCFG as a variant of PCFG. When I think of a variant, i tend to think it satisfies the independence assumption while it is not. Maybe too nitpicky, but some parsing reviewers may be}
%\Reply{I think context-free independence means two things here: from the view of CFG, C-PCFG satisfy the context-free constraint; while from the view of parameterization it is not context-free. I agree that our expression is not precise. Since PCFG is a more general model, I feel that is still ok. But I am not sure about the use of "variant" in the abstract.}
It admits tractable  estimation of the posteriors, needed in the alignment loss,
%\Ivan{Not sure marginals / posterios / partition function - but should be the same as what we refer to as a problem with Shi model. I feel that mention the partition function alone is slightly misleading}
%\Reply{Agree.}
with  dynamical programming
and leads to a fully-differentiable end-to-end visually grounded learning.
More importantly, the PCFG parser lets us  complement the alignment loss with a language modeling objective.
%offering a natural way to learn the parsing model on pure text. - I: repetetive / redundant
%\Ivan{I'd suggest to spell out the issues of the previous VG framework more clearly. See my attempt in the abstract.}
%\Reply{In the last paragraph, I tried elaborating on the issues mentioned in the abstract.}

% specific model and contributions
%$Specifically, we propose visually grounded compound PCFGs.
%It uses compound PCFGs~\citep{kim-etal-2019-compound} as the parsing model and generalizes the visually grounded grammar learning framework of~\citet{shi-etal-2019-visually}. \Ivan{The preceding two sentences can be dropped as they seem to be a bit redundant w.r.t. to the preceding text.} 
%\Ivan{I dropped two sentences here, as they were very redundant, integrated a mention of C-PCFGs in the previous paragraph.}
Our key contributions can be summarized as follows:
(1) we propose a fully-differentiable end-to-end visually grounded learning framework for grammar induction;
(2) we additionally optimize a language modeling objective to complement visually grounded learning; 
(3) we conduct experiments on MSCOCO~\citep{lin-etal-2014-mscoco} and
observe that our model has a higher recall than VG-NSL for five out of six most frequent constituent labels. For example,
it surpasses VG-NSL by 55.1\% recall on VPs and by 48.7\% recall on prepositional phrases (PPs). %and achieves an overal sentence-level F1 59.4\%.
Comparing to a model trained purely via visually grounded learning, extending the loss with a language modeling objective improves the overall F1 from 50.5\% to 59.4\%. 
%\Ivan{Minor re-occurring stylo: you cannot `jointly do smth', you can only `jointly do smth and smth.'}
%\Ivan{I feel that we can strengthen the story for combining models by saying that there are systematic differences in what is easy to learn from vision and what seems more reliably learned from LL. We can connect to analysis, and then get back to them when we talk about our own work.}
%\Ivan{I think you should talk about specific improvement. We should go beyond the story - ``we combined the losses together and got an improvement}

%\Ivan{I don't understand why you keep refering to recall. (OK, I kind of get it -- *not* producing is more direct evidence, and the improvements are great. But it is unorthodox and may confuse an (adversarial) reader, who would think that our precision may be bad.}
%\Reply{Shouldn't it be a commonsense that we can not compute precision and per-label F1 in grammar induction?}

\section{Background and Motivation}
%\Ivan{Added Motivation - to be discussed. I am worried that a reviewer who knows a lot about grammar inducton may be inclined to skip of a section called background. Currently, it is written in such way that it must be read, even if you know Shi's model.}
Our model relies on compound PCFGs~\citep{kim-etal-2019-compound} and generalizes the  visually grounded grammar learning framework of~\citet{shi-etal-2019-visually}.
We will describe the relevant aspects of both frameworks in Sections \ref{sec:cpcfg}-\ref{sec:vgnsl}, and then discuss their limitations (Section~\ref{sec:limitations}). 
%\Ivan{We really talk about limitations of grounded version, maybe still OK to say that?}
%\Reply{It is OK.}

\subsection{Compound PCFGs}\label{sec:cpcfg}

Compound PCFGs extend context-free grammars (CFGs) and, to establish notation, we start by briefly introducing them.
A CFG is defined as a 5-tuple $\mathcal{G}=(S, \mathcal{N}, \mathcal{P}, \Sigma, \mathcal{R})$
where $S$ is the start symbol, 
$\mathcal{N}$ is a finite set of nonterminals, 
$\mathcal{P}$ is a finite set of preterminals,
$\Sigma$ is a finite set of terminals,\footnote{Strictly, CFGs do not distinguish nonterminals $\mathcal{N}$ (constituent labels) from preterminals $\mathcal{P}$ (part-of-speech tags). 
	They are both treated as nonterminals. 
	$\mathcal{N}, \mathcal{P}, \Sigma$ satisfy $\mathcal{N}\cap\mathcal{P}=\emptyset$ and $(\mathcal{N}\cup\mathcal{P})\cap\Sigma=\emptyset$.}
and $\mathcal{R}$ is a set of production rules  %considered 
in the Chomsky normal form:
\begin{align*}
S\Ra A,&  &A\in \mathcal{N}, \\
A\Ra BC,&  &A\in \mathcal{N}, B, C\in\mathcal{N}\cup\mathcal{P},  \\
T\Ra w&, &T\in \mathcal{P}, w\in \Sigma\,. 
\end{align*}
PCFGs extend CPGs by associating each production rule $r\in \mathcal{R}$ with a non-negative scalar $\pi_r$ such that
$\sum_{r: A\Ra \gamma} \pi_r = 1$,
i.e., the probabilities of production rules with the same left-hand-side nonterminal sum to 1. 
%PCFGs admit exact inference and can be estimated from unlabled data with the Expectation-Maximization algorithm~\citep{EM}. 
%However, the strong context-free assumption hinders PCFGs from exploiting high-order dependencies in language. 
% I: I think the situation is a bit more nuanced as we can build strong PCFG parsers 
The strong context-free assumption hinders PCFGs and prevent them from being effective in the %natural-language 
grammar induction context.  
%\Ivan{The content above seems a bit to elementary but I do not see how to compress it and introduce the notation}\Reply{Yes, that is just a basic but serves the purpose of being self-contained, let us leave it as it is for now.}
Compound PCFGs (C-PCFGs) mitigate this issue by assuming that rule probabilities follow a compound probability distribution~\citep{robbins1951}:
\begin{flalign*}
\pi_r = g_{r}(\bz; \theta), \quad \bz\sim p({\bz}) \,, %\mathcal{N}(\mathbf{0}, \mathbf{1})\,.
\end{flalign*}
where $p(\bz)$ is a prior distribution of the latent $\bz$,
and $g_{r}(\cdot;\theta)$ is parameterized by $\theta$ and yields a rule probability $\pi_r$.
Depending on  the rule type, $g_{r}(\cdot;\theta)$  takes  one of these forms:
\begin{flalign*}
\pi_{S\Ra A} &= \frac{\exp(\bu_{A}^{T} f_s([\bw_{S}; \bz]))}
{\sum_{A'\in\mathcal{N}}\exp(\bu_{A'}^{T} f_s([\bw_{S}; \bz]))} \,, \\
\pi_{A\Ra BC} &= \frac{\exp(\bu_{BC}^{T} [\bw_{A}; \bz])}
{\sum_{B',C'\in\mathcal{N}\cup\mathcal{P}}\exp(\bu_{B'C'}^{T} [\bw_{A}; \bz])} \,, \\
\pi_{T\Ra w} &= \frac{\exp(\bu_{w}^{T} f_t([\bw_{T}; \bz]))}
{\sum_{w'\in\Sigma}\exp(\bu_{w'}^{T} f_t([\bw_{T}; \bz]))} \,,
\end{flalign*}
where $\bu$ is a parameter vector,
$\bw_{N}$ is a symbol embedding and  $N\in\{S\}\cup\mathcal{N}\cup\mathcal{P}$. 
$[\cdot;\cdot]$ indicates vector concatenation,
and $f_s(\cdot)$ and $f_t(\cdot)$ encode the input into a vector (parameters are dropped for simplicity).

A C-PCFG defines a mixture of PCFGs (i.e., we can sample a set of PCFG parameters by sampling a vector $\bz$).
It satisfies the context-free assumption conditioned on $\bz$
and thus admits exact inference for each given $\bz$.
Learning with C-PCFGs involves  maximizing the log-likelihood of every observed sentence $\boldsymbol{w}=w_1w_2\ldots w_n$:
\begin{flalign*}
%\mathcal{L}(\boldsymbol{w}; \theta) = -
\log p_{\theta}(\boldsymbol{w}) = \log
\int_{\bz} \sum_{t\in \mathcal{T}_{\mathcal{G}}(\boldsymbol{w})} p_{\theta}(t | \bz) p(\bz)\,d\bz\,,
\end{flalign*}
where $\mathcal{T}_{\mathcal{G}}(\boldsymbol{w})$ consists of all parses of the sentence $\boldsymbol{w}$ under a PCFG $\mathcal{G}$.
Though for each given $\bz$ the inner summation over parses can be efficiently computed using the inside algorithm~\citep{io_algorithm}, 
the integral over $\bz$ makes optimization intractable.
Instead, C-PCFGs rely on variational inference and maximize the evidence lower bound (ELBO):
\begin{flalign}
&\log p_{\theta}(\boldsymbol{w}) \geq \text{ELBO}(\boldsymbol{w}; \phi, \theta) =  \\
&\mathbb{E}_{q_{\phi}(\bz | \boldsymbol{w})}[\log p_{\theta}(\boldsymbol{w} | \bz)] - 
\text{KL}[q_{\phi}(\bz | \boldsymbol{w}) || p(\bz)]\,, \nonumber
\end{flalign}
where $q_{\phi}(\bz | \boldsymbol{w})$ is a variational posterior, a neural network parameterized with $\phi$.
%Intuitively, conditioning $\bz$ on $\mbs{w}$ allows us to use a sentence-specific PCFG. -> that's not really true (it is a posterior model)
The expected log-likelihood term is estimated via the reparameterization trick~\citep{kingma2014semi};
the KL term can be computed analytically when 
$p(\bz)$ and $q_{\phi}(\bz | \mbs{w})$ are normally distributed.

\subsection{Visually grounded neural syntax learner}\label{sec:vgnsl}

The visually grounded neural syntax learner (VG-NSL) comprises a parsing model and an image-text matching model.
The parsing model is an easy-first parser~\citep{goldberg-elhadad-2010-efficient}.
It builds a parse greedily in a bottom-up manner while at the same time producing a semantic representation for each constituent in the parse (i.e., its `embedding').
%Formally, given an $n$-length sentence $w_1w_2\ldots w_n$.
%The parser first encodes them into word vectors $\bw_1\bw_2\ldots\bw_n$ and 
%initializes a span set as $S_{0} = \{\bs_{1,1},\bs_{2,2},\ldots,\bs_{n,n}\}$ where $\bs_{n,n}=\bw_{n}$.
%At the $i$-th step of the construction, the parser estimates the likelihood of merging two adjacent spans in $S_{i - 1}$
%and merges the most probable pair as $\bs_{i,j} = h(\bs_{i,k}, \bs_{k+1, j})$ ($h(\cdot)$ computes the representation of the new span from the two adjacent span representations).
%Then it updates the span set as $S_{i} = S_{i - 1}\setminus\{\bs_{i,k}, \bs_{k+1, j}\}\cup\{\bs_{i,j}\}$.
%After $n-1$ steps a parse and all its constituent representations will be produced.
The parser is optimized through \textsc{Reinforce} \citep{reinforce}.
The reward  encourages merging two adjacent constituents if the merge results in a constituent that is {\it concrete}, i.e., if its semantic representations is predictive of the corresponding image, as measured with a matching function.  We omit details of the parser and how the semantic representations of constituents are computed,  as they are not relevant to our approach, and refer the reader to~\citet{shi-etal-2019-visually}.

However, as we will extend %in our work 
their image-text matching model,
we explain this component of their approach more formally. In their work, this loss  is used to learn the textual and visual representations. % and not directly affected the parser.
For every constituent $c^{(i)}$ of a sentence $\mbs{w}^{(i)}$, they define the following triplet hinge loss:
\begin{align}\label{eq:hinge_loss}
h&(\bc^{(i)},  \bv^{(i)}) =  
\mathbb{E}_{\bc'}
\left[
m(\bc', \bv^{(i)}) - m(\bc^{(i)}, \bv^{(i)}) + \epsilon
\right]_{+} \nonumber  \\
& +
\mathbb{E}_{\bv'}
\left[
m(\bc^{(i)}, \bv')\! -\! m(\bc^{(i)}, \bv^{(i)})\! +\! \epsilon
\right]_{+}
\,, 
\end{align}
where $[\cdot]_{+} = \max(0, \cdot)$,
$\epsilon$ is a positive margin,
$m(\bc, \bv) \triangleq \cos (\bc, \bv)$ is the matching function measuring similarity between the constituent representation $\bc$ and the image representation $\bv$. The expectation is taken with respect to `negative examples', $\bc'$ and $\bv'$. In practice, for efficiency reasons, a single representation of an image $\bv'$ and a single representation of a constituent (span) $\bc'$  from another example in the same batch are used as the negative examples.  
%\Ivan{Correct?}\Reply{Yes. It is exactly what I am doing.}
Intuitively, an aligned image-constituent pair $(\bc^{(i)}, \bv^{(i)})$ should score higher than an unaligned one ($(\bc', \bv^{(i)})$ or $(\bc^{(i)}, \bv')$). 
%\Ivan{I moved it as I think we should say this at the level of constituent, not the whole sentence}\Reply{I agree}

The total loss for an image-sentence pair $(\bv^{(i)}, \mbs{w}^{(i)})$ is obtained by summing losses for all constituents in a tree
$t^{(i)}$, sampled from the parsing model (we write $c^{(i)} \in t^{(i)}$):
%\begin{flalign}\label{eq:point_estimation}
%\hat{s}(\bv, \boldsymbol{w}) = \sum_{c\in t} m(\bc, \bv)\,,\,\, t \sim \texttt{PARSER}(\boldsymbol{w})\,,
%\end{flalign}
\begin{flalign}\label{eq:point_estimation}
\hat{s}(\bv^{(i)}, \boldsymbol{w}^{(i)}) = \sum_{c^{(i)}\in t^{(i)}} h(\bc^{(i)}, \bv^{(i)})\,.
\end{flalign}

In their work, training alternates between optimizing the parser using rewards (relying on image and text representations) and optimizing the image-text matching model to refine image and text representations (relying on the fixed parsing model). 
%\Ivan{I understand that they do not alternative but the point is that they have 2 somewhat separate objectives.}
%\Ivan{It may be good to clarify that there is no single objective (alternative optimization is more usual with one objectives and may even come with some guarantees; but this is - afau - is not the case ere.}
%\Reply{My description seems imprecise. Based on your explanation, their optimization does not belong to alternative optimization since they have two objectives: maximizing the rewards and minimizing the contrastive loss. Though in their paper, they are saying that they optimize them alternatively, in their code, the two objectives are optimized at the same time.}
Once trained, the parser can be directly applied to raw text, i.e., images are not used at test time.

\subsection{Limitations of the VG-NSL framework}
\label{sec:limitations}

While straightforward, % is it ?
there are several practical issues inhibiting the visually grounded learning framework.
First, contrastive learning implicitly assumes that every constituent of a sentence has its visual representation in the aligned image.
However, it is not guaranteed in practice and would result in noisy reward signals.
Besides, the loss in Equation \ref{eq:hinge_loss} (and a similar component in the reward, see \citet{shi-etal-2019-visually}) focuses on constituents corresponding to short spans.
Long spans, independently of their syntactic structure, 
tend to be sufficiently discriminative to distinguish the aligned image $\bv^{(i)}$ from an unaligned one. This implies that there is not much learning signal for such constituents.
%\Ivan{Spelled out it a bit}
The tendency to focus on short spans and those more easily derivable from an image  is evident from the results~\cite{shi-etal-2019-visually,kojima-etal-2020-learned}.
For example, their parser is accurate for noun phrases (recall 79.6\%), which are often short for captions, but performs poorly on verb phrases
(recall 26.2\%) which have longer spans, more complex compositionally and also harder to predict from images (see our analysis in Section~\ref{sec:analysis}). 
%\Ivan{Would it make sense for us to include analysis confirming this more directly, i.e. bucketed by span length? (If we have space and you have time) \textbf{Ups:} you actually did! Nice, please refer to this analysis from here.} 
%\Reply{done}
While there may be ways to mitigate some of these issues, we believe that any image-text matching loss alone is unlikely to provide sufficient learning signal to accurately captures all aspects of syntax. 
Instead of resorting to language-specific inductive biases as done by \citet{shi-etal-2019-visually} (i.e., head-initial bias~\citep{baker2008atoms} of English), we propose to complement the image-text matching loss with the objective derived from the unaligned text (i.e., log-likelihood), jointly training a parser to both explain the raw language data and the alignment with images.

%\reply{Besides, matching-based rewards can lead the parser to focus only on local short constituents such as noun phrase (NP).
%As evidenced in~\citet{shi-etal-2019-visually},
%their parser is good at NP (with an F1 79.6) 
%but performs poorly on more complex compositionality such as verb phrase (VP) (with an F1 26.2).}
%\sout{Besides, matching-based rewards can lead the parser to focus only on local noun phrase. 
%It may be incapable of learning more complex compositionality such as verb phrase.} \Ivan{Reads speculative? Can we ground it in some empirical results? Also, maybe 2 different things are mixed here: (1) the objective is informative for short constituents; (2) NP vs VP.}
%We will show that jointly training the parser on pure text can mitigate this issue.
Moreover, their learning is likely to suffer from large variance in gradient estimation as their parser does not admit tractable estimation of the partition function, 
%\Reply{this also reads speculative. I am not sure how to describe it: reinforce is likely to suffer from huge variance, but we did not observe huge variance in Shi's paper, while we know that is because their parser does not explore more complex compositionalities.}
and thus they have to rely on sampling decisions.
This will be even more of a problem if we would attempt to use it in the joint learning set-up.  Also note that similar parsing models do not yield  linguistically-plausible structures when used in the conventional (i.e., non-grounded) grammar-induction set-ups~\cite{williams-etal-2018-latent,havrylov2019cooperative}. %\Ivan{Added "also" - as it is not entirely clear if this has to do with variance.}
%\Ivan{What do you think? Is there a better citation?}\Reply{added a citation to Adina Williams's paper, they analyzed the Gumbel-Tree model}

%Finally, since the parser lack necessary expressiveness, it has to resort to language-specific inductive biases to ensure valid constituents (e.g., head-initial bias~\citep{baker2008atoms} of English).

In the next section, we will use compound PCFGs and describe an improved visually grounded learning framework that can tackle these issues neatly.

\section{Visually grounded compound PCFGs}

We use compound PCFGs~\citep{kim-etal-2019-compound} and develop visually-grounded compound PCFGs (VC-PCFGs) within the contrastive learning framework.
%Visually grounded compound PCFGs (VC-PCFGs) are built within the contrastive learning framework and use C-PCFGs as the parsing model.
Instead of sampling a tree and computing a point estimate of the image-text matching loss,
we can compute the expected image-text matching loss under a tree distribution
and use end-to-end contrastive learning (Section~\ref{sec:learning}).
Since it is inefficient to compute constituent representations
relying on the chart, 
we will introduce an additional textual representation model to encode constituents (Section~\ref{sec:representation}).
%Textual and visual representations can be optimized through the image-text matching model as in VG-NSL.
Moreover, VC-PCFGs let us additionally optimize a language modeling objective,
complementing the visually grounded contrastive learning (Section~\ref{sec:joint_objective}).

\subsection{End-to-end contrastive learning}\label{sec:learning}

In the visually grounded grammar induction framework,
the parsing model is optimized through learning signals derived from the alignment of images and constituents, as scored by 
the image-text matching model.
%It learns visual representations and constituent representations by scoring their alignments.
Denoting a set of image representations by $\mathcal{V}=\{\bv^{(i)}\}$
and the corresponding set of sentences by $\mathcal{W}=\{\boldsymbol{w}^{(i)}\}$,
the image-text matching model is optimized via contrastive learning:
\begin{flalign}\label{eq:hinge_Loss_expectation}
\mathcal{L}(\mathcal{V}, \mathcal{W}; \phi, \theta) = \sum_{i} s(\bv^{(i)}, \mbs{w}^{(i)}) \,.
\end{flalign}
We define $s(\bv^{(i)}, \mbs{w}^{(i)})$ as the loss of aligning $\bv^{(i)}$ and $\mbs{w}^{(i)}$.
In VG-NSL, it is estimated via  point estimation (see Equation~\ref{eq:point_estimation}).
While in VC-PCFGs,
given an aligned image-sentence pair $(\bv, \boldsymbol{w})$,
we compute the expected image-sentence matching loss under a tree distribution $p_{\theta}(t | \boldsymbol{w})$, leading to an end-to-end contrastive learning:
\begin{flalign}\label{eq:loss_span}
s(\bv, \boldsymbol{w}) =
\mathbb{E}_{p_{\theta}(t | \boldsymbol{w})} \sum_{c\in t} h(\bc, \mb{v})\,,
\end{flalign}
%where $c$ enumerates all constituents of a tree $t$ 
%and $m(\bc, \bv)$ computes the cosine distance between the textual and visual representations.
where $h(\bc, \bv)$ is the hinge loss of aligning the unlabeled constituent $c$ and the image $\bv$ (defined in Equation~\ref{eq:hinge_loss}).
Minimizing the hinge loss encourages an aligned image-constituent pair to rank higher than any unaligned one.  
Expanding the right-hand side of Equation~\ref{eq:loss_span}
\begin{flalign} \label{eq:full-loss}
s(\bv, \boldsymbol{w}) &=
\sum_{t\in\mathcal{T}_{\mathcal{G}}(\boldsymbol{w})} 
p_{\theta}(t | \boldsymbol{w}) \sum_{c\in t} h(\bc, \mb{v})\, \nonumber \\
&=\sum_{c\in \boldsymbol{w}} 
\underbrace{\sum_{t\in \mathcal{T}_{\mathcal{G}}(\boldsymbol{w})} \mathbb{I}_{\{c\in t\}}
	p_{\theta}(t | \boldsymbol{w})
}_{p(c | \mbs{w}): \text{ marginal of the span } c} h(\bc, \mb{v}) \nonumber \, \\
&= \sum_{c\in\boldsymbol{w}} p(c | \mbs{w}) h(\bc, \mb{v}) \,, 
\end{flalign} 
where $p(c|\mbs{w})$ is the conditional probability (i.e., marginal) of the span $c$ given $\mbs{w}$.
It can be efficiently computed with the inside algorithm and automatic differentiation~\citep{eisner-2016-inside}. 

\subsection{Span representation}\label{sec:representation}

Estimation of the expected image-text matching scores relies on span representations.
Ideally, a span representation should encode semantics of a span with its computation guided by its syntactic structure~\cite{socher-etal-2013-parsing}. 
%\Ivan{I did not quite like the combining syntax and semantics, I think this is more accurate.}
%\sout{They are disired to be able to encode both semantics and syntax of a span.} \Ivan{Pls rephrase the sentence}
%The semantic characteristics will lead to an accurate estimation of image-text matching scores.
The reliance on the predicted tree structure will result in  propagating learning signals derived from the alignment of images and sentences back to the parser. 
To realize this desideratum, 
we could follow the inside algorithm and recursively compose span representations~\citep{le2015forest,stern-etal-2017-minimal,drozdov-etal-2019-unsupervised},
which is, however, time- and memory-inefficient in practice. 

Instead, we produce span representations largely independently of the parser, as we will explain below. The only way the parser model influences this representation is through the predicted constituent label: we use its distribution to compute the representation.\footnote{Intuitively, the key learning signal for the parser in our model comes through the marginals in Equation~\ref{eq:full-loss}, not through the span representation.}

Specificially, as a trade-off for a better training efficiency,
we adopt a single-layer BiLSTM to encode spans.
A mean-pooling layer is applied over the hidden states $\mathbf{h}$ of the BiLSTM and followed by a label-specific affine transformation $f_{k}(\cdot)$ to produce a label-specific span representation $\bc_k$.
Take a span $c_{i, j} = w_i\ldots w_j$ ($0<i<j\leq n$): 
\begin{flalign}
\bc_{k} = f_{k}(\frac{1}{j - i + 1}\sum_{l = i}^{j} \mathbf{h}_l)\,.
%\mathbf{h}_i\ldots \mathbf{h}_j = \text{BiLSTM}()
%\bc = 
\end{flalign}
The BiLSTM encoding model operates at the span level and %naturally 
encodes semantics of a span.
Unlike using a single sentence-level (Bi)LSTM encoder, it guarantees that no information from words outside of the span leaks into its representations.
More importantly, it can run in $\mathcal{O}(n)$ for a sentence of length $n$ with a parallel implementation.
While the produced representation does not reflect the structural decisions made by the parser,
it can be sensitive to word order and may be affected
by its syntactic structure~\citep{blevins-etal-2018-deep}. 
%\Ivan{I think the main problem is not how syatactic it is but how much it is affected by this very parser. I think we should make it clear we understand it.}
%Though the BiLSTM model does not exlicitly encode syntax of a span,
%it is still capable of capturing shallow syntax~\citep{blevins-etal-2018-deep}. \Ivan{Do not feel this sentence fits the story. Maybe let's discuss. Overall I have been a bit struggling with this section. I can give it a shot if you like.}
%\Reply{Yes, let us discuss this section. 
%Basically, I am trying explaining why we expect a span %representation to encode both semantics and syntax, and then introduce a viable/intuitive way (based on chart parsing) to do that. But we find it impractically, so we use a BiLSTM. 
%Though the BiLSTM does not directly encode syntax, I think it is %still capable of learning some shallow syntax,
%an it might be sufficient for our task}.

In order to compute the representation of unlabeled constituent $\bc$, we average the label-specific span representation $\bc_k$ under the distribution of labels defined by the parser:
\begin{flalign}
\bc = \
\sum_{k = 1}^{K} p(k |c, \mbs{w}) \bc_k \,,
\end{flalign}
where $p( k |c, \mbs{w})$ is the probability that the span $\bc$ has label $k$, conditioned on having this constituent span in the tree.

To further reduce computation we estimate the matching loss only using the %first - unclear
$\frac{n(n - 1)}{4}$ shortest spans for a sentence of length $n$.
Thus the image-text alignment loss will focus  on small constituents.
This is the case anyway (see discussion in Section~\ref{sec:limitations}), so
we expect that this simplification would not hurt model performance significantly. 
%(verify it if possible, e.g., check score contributions by spans of different lengths).

\subsection{Joint objective}\label{sec:joint_objective}

%\sout{While the end-to-end contrastive learning can already facilitate the optimization of the parsing model, }
%\Ivan{I am not entirely sure it is true for our model.}
%\Reply{No, it is not ture.}
Rather than simply optimizing the contrastive learning objective,  
we additionally maximize the log-likelihood of text data.
As with C-PCFGs, we optimize the ELBO:
\begin{flalign}
\mathcal{L}(\mathcal{W}; \phi,\theta) = -\sum_{\boldsymbol{w}\in\mathcal{W}}\text{ELBO}(\boldsymbol{w}; \phi, \theta)\,.
\end{flalign}
This learning objective complements contrastive learning.
As contrastive learning optimizes a parser by solely matching images and constituents,
the parser would only focus on simple and local constituents (e.g., short NPs). 
% not all NPs are simple actually 
Moreover, in practice, since not every constituent can be grounded in an image,
contrastive learning would suffer from misleading or ambiguous learning signals.
%As we will show, \Ivan{It may be read as we do the ablation within our model, we do not. Drop?}
%incorporating the language modeling objective can mitigate these issues,
%and lets the parser  be accurate across phrase-structure types. 
%\Ivan{We have a bit of redundancy across the sections.}

To summarize, the overall loss function is
\begin{flalign}
\!\!\!\mathcal{J}(\phi,\theta) = \mathcal{L}(\mathcal{W}; \phi,\theta) + \alpha\cdot\mathcal{L}(\mathcal{V}&, \mathcal{W}; \phi, \theta)\,,
\end{flalign}
where $\alpha$ is a hyper-parameter balancing the relative importance of the contrastive learning.

\subsection{Parsing}

The parser can be directly used to parse raw text after training,
without requiring access to visual groundings.
Parsing seeks for the most probable parse $t^{*}$ of $\boldsymbol{w}$:
\begin{align*}
t^{*} = \argmax\int_{\bz} p_{\theta}(t | \bw, \bz) p_{\theta}(\bz | \boldsymbol{w})\,d\bz\,.
\end{align*}
Still, though the maximum a posterior (MAP) inference over $p_{\theta}(t | \boldsymbol{w})$ can be solved by the CYK algorithm~\citep{kasami1966efficient,YOUNGER1967189},
inference becomes intractable when introducing into $\bz$.
The MAP inference is instead approximated by
\begin{flalign*}
t^{*} \approx \argmax\int_{\bz} p_{\theta}(t | \bw, \bz) \delta(\bz - \bmu_{\phi}(\boldsymbol{w}))\,d\bz\,,
\end{flalign*}
where $\delta(\cdot)$ is the Dirac delta function
and $\bmu_{\phi}(\boldsymbol{w})$ is the mean vector of the variational posterior $q_{\phi}(\bz | \boldsymbol{w})$.
As $\delta(\cdot)$ has zero mass everywhere but at the mode $\bmu_{\phi}(\boldsymbol{w})$,
it is equivalently solving $\argmax_{t}p_{\theta}(t | \boldsymbol{w}, \bmu_{\phi}(\boldsymbol{w}))$.

\section{Experiments}

\subsection{Datasets and evaluation}\label{sec:data_and_eval}

\noindent{\textbf{Datasets}}:
We use MSCOCO~\citep{lin-etal-2014-mscoco}.
It consists of 82,783 training images, 1,000 validation images, and 1,000 test images.
Each image is associated with 5 caption sentences.
We encode images into 2048-dimensional vectors using the pre-trained ResNet-101~\citep{he2016deep}.
At test time, only captions are used.
We follow~\citet{shi-etal-2019-visually} and parse test captions with Benepar~\citep{kitaev-klein-2018-constituency}.
We use the same data preprocessing\footnote{\url{https://git.io/JfV6J}.} as in~\citet{shen2018ordered} and~\citet{kim-etal-2019-compound},
where punctuation is removed from all data,
and the top 10,000 frequent words in training sentences are kept as the vocabulary.
%
%\citet{shi-etal-2019-visually} \sout{also use the most frequent 10,000 words in the training captions as the vocabulary, but they keep punctuation.} \Ivan{Having the last sentence looks slightly strange. Maybe say why we do differently. Also can be moved to a footnote? }\Reply{will mention this in the results section, I intended to emphasize on the difference in dealing with punctuations.}

\noindent{\textbf{Evaluation}}:
We mainly compare VC-PCFGs with VG-NSL~\citep{shi-etal-2019-visually}.
To verify the effectiveness of the use of visual groundings,
we also compare our model with a C-PCFG trained only on the training captions.
All models are run four times with different random seeds and for at most 15 epochs with early stopping (i.e., the image-caption loss / perplexity on the validation captions does not decrease).
We report both averaged corpus-level F1 and averaged sentence-level F1 numbers as well as the unbiased standard deviations.
%discard trivial spans such as single-word spans and sentence-level spans at test.
%As a typical practice, punctuation in the test captions are ignored~\citep{shen2018neural,shen2018ordered,kim-etal-2019-compound}.

\subsection{Settings and hyperparameters}

We adopt parameter settings suggested by the  authors for the baseline models.
For VG-NSL we run the authors'
%\Ivan{I am not sure we should use these compressed urls. Are they even temporary? }\Reply{Maybe that is fine? The tiny url service is provided by Github Inc, so are the git repositories. I think these urls will be always valid before Github is closed. I use them mostly for a clean presentation. :)}
code.\footnote{\url{https://git.io/Jf3nn}.}
We re-implement C-PCFG using automatic differentiation~\citep{eisner-2016-inside} to speed up training.
Our VC-PCFG comprises a parsing model and an image-text matching model.
The parsing model has the same parameters as the baseline C-PCFG;
the image-text matching model has the same parameters as the baseline VG-NSL.
Concretely,
the parsing model has 30 nonterminals and 60 preterminals.
Each of them is represented by a 256-dimensional vector.
The inference model $q_{\phi}(\mb{z} | \boldsymbol{w})$ uses a single-layer BiLSTM.
It has a 512-dimensitional hidden state and relies on 512-dimensitional word embeddings.
We apply a max-pooling layer over the hidden states of the BiLSTM 
and then obtain 64-dimensitional mean vectors $\bmu_{\phi}(\boldsymbol{w})$ and log-variances $\log\mb{\sigma}_{\phi}(\boldsymbol{w})$ by using an affine layer.
The image-text matching model projects visual features into 512-dimensitional feature vectors and encodes spans as 512-dimensitional vectors.
Our span representation model is another single-layer BiLSTM,
with the same hyperparameters as in the inference model.
$\alpha$ for visually grounded learning is set to 0.001.
We implement VC-PCFG relying on  Torch-Struct~\citep{rush-2020-torch},
and optimize it using Adam~\citep{kingma2014adam} with the learning rate set to 0.01, $\beta_1 = 0.75$, and $\beta_2=0.999$.
All parameters are initialized with Xavier uniform initializer~\citep{pmlr-v9-glorot10a}. 
%\Ivan{Some of these can go into appendix.}

\subsection{Results and analysis}

\subsubsection{Main results}

\begin{table*}[t]\small
	\centering
	{\setlength{\tabcolsep}{.65em}
		\makebox[\linewidth]{\resizebox{\linewidth}{!}{%
				\begin{tabular}{rllllllll}
					\toprule
					\multicolumn{1}{r}{\textbf{Model}} &
					\multicolumn{1}{l}{\textbf{NP}} & 
					\multicolumn{1}{l}{\textbf{VP}} & 
					\multicolumn{1}{l}{\textbf{PP}} & 
					\multicolumn{1}{l}{\textbf{SBAR}} & 
					\multicolumn{1}{l}{\textbf{ADJP}} & 
					\multicolumn{1}{l}{\textbf{ADVP}} & 
					\multicolumn{1}{l}{\textbf{C-F1}} & 
					\multicolumn{1}{l}{\textbf{S-F1}} \\
					\midrule
					Left Branching & 33.2 & \phantom{0}0.0 & \phantom{0}0.1 & \phantom{0}0.0 & \phantom{0}4.9 & \phantom{0}0.0 & 15.1 & 15.7 \\
					%Right Branching & 23.8 & 91.5 & 63.0 & \textbf{96.0} & 18.3 & 76.7 & 42.4 & 42.8 \\
					Right Branching & 37.5 & \textbf{94.5} & 71.1 & \textbf{97.8} & 20.9 & 79.1 & 51.0 & 51.8 \\
					Random Trees & 32.8$_{\pm 0.5}$ & 18.4$_{\pm 0.4}$ & 24.4$_{\pm 0.3}$ & 17.7$_{\pm 1.7}$  & 26.8$_{\pm 2.6}$ & 20.9$_{\pm 1.5}$ & 24.2$_{\pm 0.3}$ & 24.6$_{\pm 0.2}$ \\
					%		PRPN & \\
					%		ON-LSTM & \\
					C-PCFG &  43.0$_{\pm 8.6}$ & 85.0$_{\pm 2.6}$ & 78.4$_{\pm 5.6}$& 90.6$_{\pm 2.1}$& 36.6$_{\pm 21}$& \textbf{87.4}$_{\pm 1.0}$& 53.6$_{\pm 4.7}$& 53.7$_{\pm 4.6}$\\
					\midrule
					VG-NSL$^\dagger$ & 79.6$_{\pm 0.4}$ & 26.2$_{\pm 0.4}$ & 42.0$_{\pm 0.6}$ & & 22.0$_{\pm 0.4}$ &  & 50.4$_{\pm 0.3}$ & \\
					VG-NSL+HI$^\dagger$ & 74.6$_{\pm 0.5}$ & 32.5$_{\pm 1.5}$ & 66.5$_{\pm 1.2}$ & & 21.7$_{\pm 1.1}$  & & 53.3$_{\pm 0.2}$ &  \\
					%		\midrule
					VG-NSL$^\star$ & \textbf{64.3}$_{\pm 1.1}$ & 28.1$_{\pm 0.5}$ & 32.2$_{\pm 1.1}$ & 16.9$_{\pm 3.2}$ & 13.2$_{\pm 1.5}$  & \phantom{0}5.6$_{\pm 0.3}$  & 41.5$_{\pm 0.5}$ & 41.8$_{\pm 0.5}$ \\
					VG-NSL+HI$^\star$ & 61.0$_{\pm 0.2}$ & 33.5$_{\pm 1.6}$ & 62.7$_{\pm 0.6}$ & 42.0$_{\pm 5.1}$ & 13.9$_{\pm 0.6}$ & 65.9$_{\pm 2.5}$ & 48.8$_{\pm 0.4}$ & 49.4$_{\pm 0.5}$   \\
					\midrule
					VC-PCFG (ours) &  54.9$_{\pm 14}$& 83.2$_{\pm 3.9}$& \textbf{80.9}$_{\pm 7.9}$& 89.0$_{\pm 2.0}$& \textbf{38.8}$_{\pm 25}$& 86.3$_{\pm 4.1}$& \textbf{59.3}$_{\pm 8.2}$& \textbf{59.4}$_{\pm 8.3}$\\
					w/o LM &  35.6$_{\pm 3.7}$ &  {93.4}$_{\pm 2.1}$ &  70.1$_{\pm 2.0}$ &  95.9$_{\pm 3.9}$ &  20.6$_{\pm 0.8}$ &  78.0$_{\pm 2.2}$ &  49.7$_{\pm 2.6}$ &  50.5$_{\pm 2.5}$ \\
					\bottomrule
	\end{tabular}}}}
	\caption{\label{tab:eval_mscoco}
		Recall on six frequent constituent labels (NP, VP, PP, SBAR, ADJP, ADVP) in the MSCOCO test captions and corpus-level F1 (C-F1) and sentence-level F1 (S-F1) results.
		The best mean number in each column is in bold.
		$\dagger$ indicates results reported by~\citet{shi-etal-2019-visually}.
		$\star$ denotes results obtained by running their code.
		Notice that the results from~\citet{shi-etal-2019-visually} are not comparable to ours
		because they keep punctuation and include trivial sentence-level spans in evaluation.
	}
\end{table*}

Our model outperforms all baselines according to both corpus-level F1 and sentence-level F1 (see  Table~\ref{tab:eval_mscoco}). 
Notably, it surpasses VG-NSL+HI by 10\% F1.\footnote{
	We run the code of~\citet{shi-etal-2019-visually} and train VG-NSL and VG-NSL+HI on the training captions with punctuation removed.
	This is considered  a more challenging setting as punctuation signals the boundaries of constituents and makes it easy for parsers to derive constituents.
	At test time, as a common practice~\citep{shen2018neural,shen2018ordered,kim-etal-2019-compound}, 
	we discard punctuation and ignore trivial single-word and sentence-level spans. 
	We notice that including sentence-level spans can improve the F1 of VG-NSL to around 49\%.
}	
%\Ivan{Maybe repeat what trivial spans are and this is the comming practive in other work? You can remove it from other section then, if short of space.} 
%without relying on  the head-initial (HI) property of English~\citep{baker2008atoms} used in VG-NSL+HI. \Ivan{Small comment -- you tend to make English as over-regular ;)  words `punctuations', `compositionalities', `accuratest', `recalls' do not exist.}
The right branching model is a strong baseline on image captions, as observed previously on the
WSJ corpus,
%Wall Street Journal portion of the Penn Treebank
%~\citep{marcus-etal-1994-penn} is observed
including in recent work~\citep{shen2018neural,kim-etal-2019-compound}. 
%\Ivan{Rephrased, as (1) sounded like it is a new observation; (2) Mitch Marcus et al will survive without yet another citation.} 
Comparing with C-PCFG, which is trained solely on captions,
VC-PCFG achieves a much higher mean F1 (+5.7\% F1),
%\Ivan{Please check} 
demonstrating the informativeness of visual groundings.
However, VC-PCFG suffers from a larger variance
%\Ivan{Not sure if should keep this footnote about one outlayer.}\Reply{deleted as it is unclear where the variance comes from}
%\footnote{We notice that the huge variance is caused by one unlucky run of our model.}
%because the matching-based contrastive learning inevitably is afflicted by misleading learning signals (see discussion in Section~\ref{sec:limitations}). 
presumably because the joint objective is harder to optimize.
%\Ivan{Not sure how this fits the story we wrote (I think we were blaming reinforce for the variance; otherwise, I think, it is bias so not clear why increases the variance)}
%\Reply{I did not intend to blame reinfore for the variance. I think the variance is caused by contrastive learning's implicitly assumption that every constituent has its visual correspondence in the aligned image} 
%\Ivan{Not sure why it would cause variance / instability (rather than bias). Also VG-NSL is pretty stable. Suggestion: maybe say 'presumably because the joint objective is harder to optimize' (I think it is a more likely reason) }
%Visually grounded contrastive learning (w/o LM) shows a moderate performance,
Visually grounded contrastive learning (w/o LM) has a mean F1 50.5\%.
It is further improved to 59.4\% when additionally optimizing the language modeling objective.
%\Ivan{a moderate performance -- needs to be changed (not fluent) but not sure to what as there is no result yet}
%When jointly optimizing the language modeling objective, the mean F1 is improved from 50.5\% to 59.4\%. 
%\Ivan{If you don't mind add \% after all numbers. I think it is a std convention not to drop them.}

Moreover, we show recall on six frequent constituent labels (NP, VP, PP, SBAR, ADJP, ADVP) in the test captions.
Unsurprisingly, VG-NSL is best on NPs because the matching-based reward signals optimize it to focus only on short and concrete NPs (recall 64.3\%). 
It performs poorly on other constituent labels such as VPs (recall 28.1\%).
%that involve more complex compositionalities.  - no such word compositionality, I also could imagine someone disagreing that VP is more complex in general (rather with controversial PTB notation which does not annotate internal NP structure)
%Though the head-initial inductive bias improves VG-NSL on VPs and PPs,
%it is still far behind the linguisticly motivated C-PCFG parser overally.
In contrast, VC-PCFG exhibits a relatively even performance across constituent labels,
e.g., it is most accurate on SBARs and ADVPs and works fairly well on VPs (recall 83.2\%). 
Meanwhile, it improves over C-PCFG for NPs, which are usually short and `concrete',
once again confirming the benefits of using visual groundings.
Visually grounded contrastive learning (w/o LM) tends to behave like the right branching baseline.
%while performing slightly better on NPs (+2.8\% recall).
Additionally optimizing the language modeling objective brings a huge improvement for NPs (+19.3\% recall).
%Compared with C-PCFG, visual groundings bring a huge improvement on NPs
%because contrastive learning encourages the parser to focus on short and concrete NPs.
%\Ivan{Try to repharse it to show that our model gains from grounding, not only improves over VG-NSL. We need to show comlementarity of both signals better.}

\begin{figure}
	\begin{center}
		\includegraphics[width=1.\linewidth]{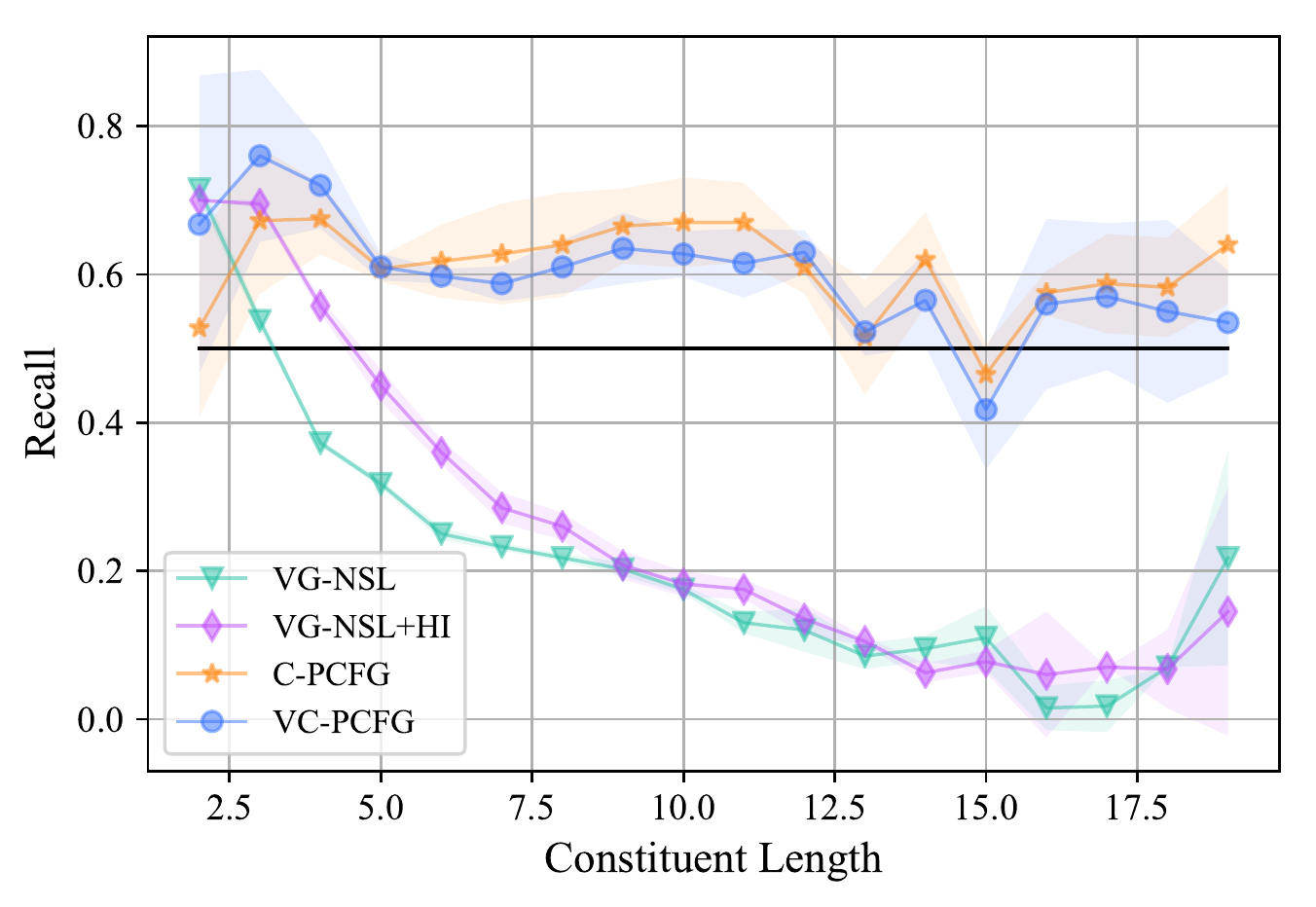}
	\end{center}
	\caption{\label{fig:recall_per_length}
		Recall broken down by constituent length. %\reply{todo: readable image with no color.}
		%		\ivan{(1) One point which this plot doesn't reveal that short constituents are the most frequent ones. Is there some way to represent it? 
		%		(2) red and purple are too similar. (3) Any chance to make it readable with no color (e.g., using dashes)? - secondary improtance for now.}
	}
\end{figure}

\subsubsection{Analysis}\label{sec:analysis}

\begin{figure}
	\begin{center}
		\includegraphics[width=.75\linewidth]{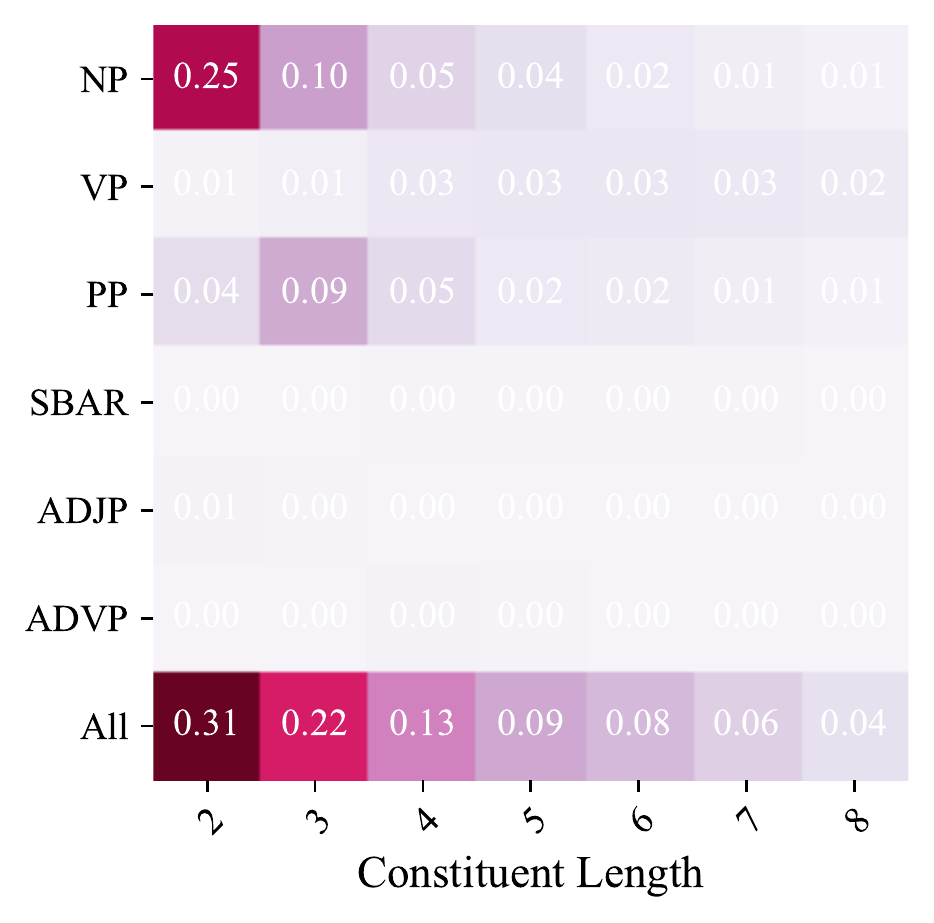}
	\end{center}
	\caption{\label{fig:label_dist_per_length}
		Label distribution over constituent length. 
		\textit{All} denotes frequencies of constituent lengths.
		Zero frequencies are due to the limited numerical precision.
	}
\end{figure}

We analyze model performance for constituents of different lengths (Figure~\ref{fig:recall_per_length}).
As expected, VG-NSL becomes weaker as constituent length increases, and the drop is very dramatic. 
%All models tend to be good at longest spans because these spans are similar to trivial sentence-level spans.
C-PCFG and its grounded version VC-PCFG consistently outperform VG-NSL on constituents longer than four tokens
and display a more even performance across constituent lengths.
Meanwhile, VC-PCFG  beats C-PCFG on constituents of length below 5,
confirming that visual groundings are beneficial for short spans.
We further plot the distribution over constituent length for different phrase types (Figure~\ref{fig:label_dist_per_length}) and find that around 75\% constituents in our dataset are shorter than six tokens,
and 60\% of them are NPs.
Thus, it is not surprising that the improvement on NPs, brought by  visually grounded learning, has a large impact on the overall performance.

Next, we analyze induced tree structures.
We compare model predictions against gold trees,  left branching trees, and right branching trees.
As there is little performance difference between corpus-level F1 and sentence-level F1, 
we focus on sentence-level F1 in this analysis.
We report self F1~\citep{williams-etal-2018-latent} to show model consistency across runs.
The self F1 is computed by averaging over six model pairs from four different runs.
All results are presented in Table~\ref{tab:eval_mscoco_sim}.
Overall, all models have self F1 above 70\%, indicating a relatively high consistency.
We observe that using the head-initial bias pushes VG-NSL closer to the right-branching baseline,
while visual grounded learning leads to improvements over C-CPFG, forcing VC-PCFG to deviate
from the default right-branching behaviour. 

\begin{figure}
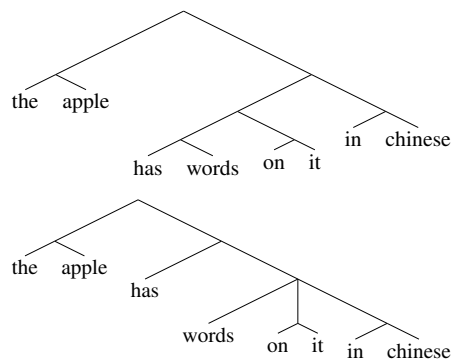

	\centering
	\resizebox{.75\linewidth}{!}{%
		\Tree [[the apple ] [[[has words ] [on it ] ] [in chinese ] ] ] } \\
	\vspace{.5em}
	\resizebox{.75\linewidth}{!}{%
		\Tree [[the apple ] [has [words [on it ] [in chinese ] ] ] ]}
	\caption{\label{fig:sample_trees}
		Upper: A parse output by the best run of VC-PCFG. Bottom: The corresponding gold tree.
	}
\end{figure}

Finally, we test VG-NSL+HI and VC-PCFG on 50 manually annotated captions released by~\citet{shi-etal-2019-visually}.
VC-PCFG achieves a mean F1 62.7\%, %\phantom{$_{\pm 8.8}$}, 
surpassing VG-NSL+HI by 12.1\% F1.
In Figure~\ref{fig:sample_trees} we visualize a parse tree predicted by the best run of VC-PCFG. 
We can see that VC-PCFG identifies most NPs but makes mistakes in PP attachement and consequently fails to identify the VP.
%(correct and incorrect predictions) - not clear

\begin{table}[t]\small
	\centering
	{\setlength{\tabcolsep}{.35em}
		\makebox[\linewidth]{\resizebox{\linewidth}{!}{%
				\begin{tabular}{rllll}
					\toprule
					\textbf{Model} & 
					\multicolumn{1}{l}{\textbf{Gold}} & 
					\multicolumn{1}{l}{\textbf{Left}} & 
					\multicolumn{1}{l}{\textbf{Right}} & 
					\multicolumn{1}{l}{\textbf{Self}}   \\
					\midrule
					%		VG-NSL & 41.8$_{\pm 0.5}$ & 28.3$_{\pm 0.5}$ & 20.6$_{\pm 0.3}$  & 84.3$_{\pm 0.4}$  \\
					%		VG-NSL+HI & 49.4$_{\pm 0.5}$ & 24.5$_{\pm 0.5}$  & 29.2$_{\pm 0.6}$  & 88.6$_{\pm 0.8}$   \\
					%		C-PCFG & 53.7$_{\pm 4.6}$ & \phantom{0}1.3$_{\pm 0.8}$  & 53.6$_{\pm 1.3}$  & 77.3$_{\pm 5.6}$   \\
					%		VC-PCFG & \textbf{59.4}$_{\pm 8.3}$ & \phantom{0}4.4$_{\pm 4.7}$  & 48.5$_{\pm 6.8}$  & 71.1$_{\pm 6.0}$   \\
					VG-NSL & 41.8\phantom{$_{\pm 0.5}$} & 28.3\phantom{$_{\pm 0.5}$} & 20.6\phantom{$_{\pm 0.3}$}  & 84.3\phantom{$_{\pm 0.4}$}  \\
					VG-NSL+HI & 49.4\phantom{$_{\pm 0.5}$} & 24.5\phantom{$_{\pm 0.5}$}  & 29.2\phantom{$_{\pm 0.6}$}  & 88.6\phantom{$_{\pm 0.8}$}   \\
					C-PCFG & 53.7\phantom{$_{\pm 4.6}$} & \phantom{0}1.3\phantom{$_{\pm 0.8}$}  & 53.6\phantom{$_{\pm 1.3}$}  & 77.3\phantom{$_{\pm 5.6}$}   \\
					VC-PCFG & 59.4\phantom{$_{\pm 8.3}$} & \phantom{0}4.4\phantom{$_{\pm 4.7}$}  & 48.5\phantom{$_{\pm 6.8}$}  & 71.1\phantom{$_{\pm 6.0}$}   \\
					\bottomrule
	\end{tabular}}}}
	\caption{\label{tab:eval_mscoco_sim}
		Average sentence-level F1 results against gold trees (Gold), left branching trees (Left), right branching trees (Right), and self F1 (Self)~\citep{williams-etal-2018-latent}.
	}
\end{table}

\section{Related work}

\textbf{Grammar Induction} has a long history in computational linguistics. Following  observations that direct optimization of log-likelihood with the Expectation Maximization algorithm~\cite{LARI199035} is not effective at producing effective grammars, a number of approaches have been developed, emboding various inductive biases or assumption about the language structure and its relation to surface realizations~\cite{klein-manning-2002-generative,smith2005guiding,cohen2009shared,spitkovsky2010viterbi}. The recent advances in the area have been brought by flexible neural models~\cite{jin-etal-2019-unsupervised,kim-etal-2019-compound,kim-etal-2019-unsupervised,drozdov-etal-2019-unsupervised}. All these methods, with the exception of \citet{shi-etal-2019-visually}, rely solely on text.

\textbf{Visually grounded learning} is motivated by the observation that natural language is grounded in perceptual experiences~\citep{steels1998,barsalou_1999,fincher-2001,ROY2002353,bisk2020experience}.
It has been shown effective in word representation learning~\citep{bruni-2014,silberer-lapata-2014-learning,lazaridou-etal-2015-hubness} and sentence representation learning~\citep{kiela-etal-2018-learning,bordes-etal-2019-incorporating}.
All this work uses visual images as perceptual experience of language 
and exploits visual semantics derived from images to improve continuous vector representatios of language.
%While we consider a more challenging task, 
In contrast, we induce structured representations, discrete tree structure of language, by using visual groundings.
We propose a model for the task within the contrastive learning framework.
Learning involves estimating \textit{concreteness} of spans,
which generalizes word-level concreteness~\citep{turney-etal-2011-literal,kiela-etal-2014-improving}.

In the vision and machine learning community,  unsupervised \textbf{induction of structured image representations} (aka scene graphs or world models) has been receiving increasing attention~\cite{eslami2016attend,burgess2019monet,kipf2019contrastive}. However, they typically rely solely on visual signal. An interesting extension of our work would be to consider joint induction of structured representations of images and text while guiding learning by an alignment loss.
%\Ivan{If you decide to drop this paragraph, I don't mind.}

%There is very little work which tries to bridge the modalities in fully unsupervised fashion, and, with the exception of \citet{shi-etal-2019-visually}, these consider on semantic rather than syntactic relations on the language side (e.g., \cite{can-we-cite-smth-here}).

\section{Conclusion}

We have presented visually-grounded compound PCFGs (VC-PCFGs) that use compound PCFGs and generalize the visually grounded grammar learning framework.
VC-PCFGs exploit visual groundings via contrastive learning, with
learning signals  derived from minimizing an image-text alignment loss.
To tackle the issues of misleading and insufficient learning signals from purely agreement-based learning,
we propose to complement the image-text alignment loss with a loss defined on unlabeled text.
We resort to using compound PCFGs which enables us to complement the alignment loss with a language modeling objective, resulting in a fully-differentiable end-to-end visually grounded learning.
We empirically show that our VC-PCFGs are superior to models that are trained only through visually grounded learning or only relying on text.

%VC-PCFGs enjoy relatively fast training with a parallel implementation of the dynamical programming.
%It takes around eight hours to iterate over 410k sentences on a single NVIDIA GTX 1080 GPU.
%However, due to the high space complexity ($\mathcal{O}(n^2)$) of the dynamical programming and limited GPU memory, 
%we can only use a small batch size of 5.
%%Thus our model potentially suffers from variance of gradient estimation.
%Moreover, we observe a higher variance from VC-PCFG,
%though compound PCFGs themselves are plagued by model variance as well (also observed by~\citet{kim-etal-2019-compound}), \Ivan{Should go to appendix, very strange to see it in conclusions. }
%
%As our focus is on a generalized visually-grounded grammar induction framework,
%we did not tune model hyperparameters.
%Since C-PCFGs are tuned on the WSJ corpus, a different domain from image captions,
%we may expect a better performance by tuning VC-PCGFGs on MSCOCO captions. \Ivan{I don't think we want to emphasize this aspect. Someone can say that we should have tune to make findings reliable, let's not make it easy for him }
\section*{Acknowledgments}
We would like to thank anonymous reviewers for their 
suggestions and comments. The project was supported by the
European Research Council (ERC Starting Grant BroadSem
678254) and the Dutch National Science Foundation
(NWO VIDI 639.022.518).

\bibliographystyle{acl_natbib}
\bibliography{emnlp2020}

\end{document}